\documentclass[10pt,twocolumn,letterpaper]{article}

\usepackage{3dv}
\usepackage{times}
\usepackage{epsfig}
\usepackage{graphicx}
\usepackage{amsmath}
\usepackage{amssymb}
\usepackage{adjustbox}
\usepackage{subcaption}

\usepackage{xcolor}
\usepackage{soul}

\usepackage[pagebackref=true,breaklinks=true,letterpaper=true,colorlinks,bookmarks=false]{hyperref}

\threedvfinalcopy  %*** Uncomment this line for the final submission

\def\threedvPaperID{43} % *** Enter the 3DV Paper ID here

\begin{document}

%%%%%%%%% TITLE
\title{Best Viewpoint Tracking for Camera Mounted \\on robot Arm with Dynamic Obstacles}

\author{Christos Maniatis \and Marcelo Saval-Calvo* \and Radim Tyle{\v c}ek \and Robert B. Fisher\\
School of Informatics, University of Edinburgh, \texttt{cmaniatis,rtylecek,rbf@inf.ed.ac.uk}.\\
*Dept. Computer Technology, University of Alicante, \texttt{msaval@dtic.ua.es}.
%{\tt\small secondauthor@i2.org}
}

\maketitle
\thispagestyle{empty}

%%%%%%%%% ABSTRACT
\begin{abstract}
The problem of finding a next best viewpoint for 3D modeling or scene mapping has been explored in computer vision over the last decade. 
This paper tackles a similar problem, but with different characteristics.
It proposes a method for dynamic next best viewpoint recovery of a target point while avoiding possible occlusions. 
Since the environment can change, the method has to iteratively find the next best view with a global understanding of the free and occupied parts. 

We model the problem as a set of possible viewpoints which correspond to the centers of the facets of a virtual tessellated hemisphere covering the scene. 
Taking into account occlusions, distances between current and future viewpoints, quality of the viewpoint and joint constraints (robot arm joint distances or limits), we evaluate the next best viewpoint. 
The proposal has been evaluated on 8 different scenarios with different occlusions and a short 3D video sequence to validate its dynamic performance.  
\end{abstract}

%%%%%%%%% BODY TEXT
\section{Introduction}\label{sec:intro}
There are multiple scenarios in computer vision when it is necessary to find a proper location for a sensor in dynamic environments to perceive a target point such as assembly lines, surveillance, medical operation recording \cite{Gambadauro2012}. For example, one might want to have a video record of an operation, e.g. to demonstrate that the doctor followed best practice. Given the changing positions of the doctors' hands, limbs and tools, the relative position of the camera will also need to change dynamically to maximize viewability and also avoid occlusions. 

In this paper we will focus on dynamic viewpoint selection with a moving sensor along with static and dynamic obstacles occluding a target point. The paper assumes a camera held by a robot in the configuration seen in Fig.~\ref{fig:setup}. A similar problem is the Next Best View (NBV) \cite{Massios1998,Sanchiz1999} where authors propose a method for planning of the camera positions to perform a 3D reconstruction of a given object. The problem addressed here considers how to maintain an optimal viewpoint, as contrasted with the traditional NBV problem, which aims to maximize total viewing. The optimality condition includes both visibility and camera motion terms. In our approach we will use the idea of the voxel map described in these papers that address NBV in order to reconstruct the scene from a set of 3D points. 

This paper proposes a novel method for tracking the best viewpoint based on a 3D representation of the scene. The most important aspect of the algorithm is to handle both dynamic and static elements that occlude the target point. Our algorithm has to estimate the next best viewpoint and to achieve that a partitioning of the space is proposed as a tessellated hemisphere, where the center of every face is a possible camera position or viewpoint (hereafter called `view'). The adequacy of each position in terms of occlusion, angle of perception and distance from the previous position will be defined using an objective function.  

The main contributions of this paper are twofold:
\begin{itemize}
\item Introduction of a new type of problem: the Dynamic Next Best View (DNBV), where dynamic occlusions need to be overcome.
\item An algorithm that solves the problem of DNBV.
\end{itemize}

In the following sections we will give an overview of the related work and introduce a description of the observed scene. Subsequently we will describe our optimization based solution and show experiments on real data.

% #######################################################
\section{Related work}\label{sec:soa}
The problem considered here is to find the best positions of the sensor to perceive a target point in a dynamic environment. We assume that the target to be observed remains stationary, but there are moving objects that occlude the target from different viewpoints at different times. Hence, there are two main aspects to take into account: best view selection and tracking of the target location.

%------ NBV
Similar problems related to the former aspect have been tackled. Finding the best viewpoint was coined in the literature as Next Best View (NBV) \cite{Morooka1998}. The NBV problem arises in the construction of a 3D model of an object. The aim is to find the best positions of the camera to perceive all the parts of the object \cite{Massios1998, Sanchiz1999,Pito1999,Karaszewski2016}.

Morooka et al. \cite{Morooka1998} proposed an on-line algorithm that chooses the NBV based on an already obtained partial model. It uses an objective function that takes into account the possibility of merging new data, local shape changes, control distribution and registration accuracy. In \cite{Sanchiz1999} the NBV is defined with a camera pose which simultaneously allows good registration, elimination of occlusion plane areas and observation of unseen areas. 

The problem of NBV is still a challenging problem to reduce the number of views needed to capture the whole object. Singh et al. \cite{Singh2015} uses a labeling of the object in levels of perception quality given by the angle between camera and the object and ray tracing techniques. With this information, mean-shift clustering is applied and the cluster with the highest value is chosen as the next best view. Other approaches use contours to calculate the unseen parts \cite{Monica2017}. Vasquez-Gomez et al. \cite{Vasquez-Gomez2014} used a two stage system that improves the quality of the modeling by predicting a next-best-view and evaluating a set of neighbor views, eventually selecting the best among all of them.

Our problem includes dynamic elements in the scenario, so we have to deal with moving occlusions. Zhang et al. \cite{Zhang2017} use the information of the occluding element and assigns the next best view by maximizing the perception of the surface out of the occluded region. This solution is not applicable to our problem since our environment is dynamic so the occlusion changes over time. 

%------ tracking
A second important aspect is to track the best viewpoint. Since the environment is dynamic, new occlusions can appear. Moreover, we are in a real environment where the camera movement has to be taken into account. We want to minimize the distance of displacement between viewpoints to increase image stability while maximizing the view quality as well as avoiding occlusions.

The dynamic visual tracking problem is a completely different problem to NBV. A camera mounted on a moving or stationary base is used to track an object. In \cite{Papanikolopoulos1993} the authors defined visual tracking in 2D of a single feature point as the translation $(T_{x},T{y})$ and rotation $(R_{z})$ with respect to the camera frame that keeps $\Omega_{w}$ stationary, where $\Omega_{w}$ is the area in the image plane where the target is projected. Many authors e.g. \cite{Papanikolopoulos1993,Hunt1982} have addressed this problem using different approaches.

Papanikolopoulos et al.~\cite{Papanikolopoulos1993} combined visual tracking and control. In particular they use sum of square differences of the optical flow to compute the discrete displacement which is fed to either a PI/pole assigned controller or a Kalman filter to improve the state estimates. These estimates are then used to move the robot arm. According to \cite{Papanikolopoulos1993}, \cite{Hunt1982} predicted mathematically the position of the object's centroid, in order to visually track the object. Their algorithm could only work for slowly moving objects, since it had to compute the coordinates of the centroid. 

In our case, we are not focused on tracking the occlusions themselves. Our goal is to track the free views. Some authors have studied planning methods to achieve best camera positions, also called camera planning. Dunn et al. \cite{Dunn2009} presented a proposal with a recursive path planner to perform NBV. However, the environment does not change in their case, whereas we have to re-estimate the new position in terms of new occlusions, distance between current and future point and viewpoint quality. 

Our problem to dynamically find the next best view includes multiple variables: the view quality, current occlusions in a dynamic environment and distance minimization between the current viewpoint and the new one. To the best of our knowledge, there is no previous work combining these problems. Our paper presents a novel dynamic next best view camera planning algorithm including these three main variables.

% #######################################################
\section{Proposed solution}\label{sec:prop-sol}

\begin{figure*}[htb]
  \includegraphics[width=\textwidth]{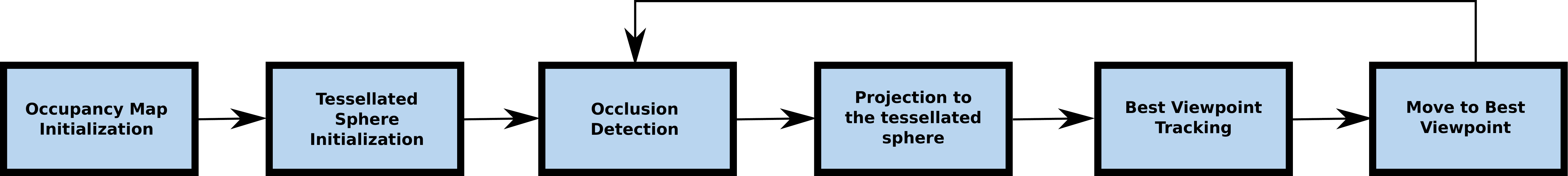}
  \centering
  \vspace{2mm}
  \caption{Scheme of the proposed method.}
  \label{fig:scheme}
\end{figure*}

\begin{figure}
  \includegraphics[width=0.9\linewidth]{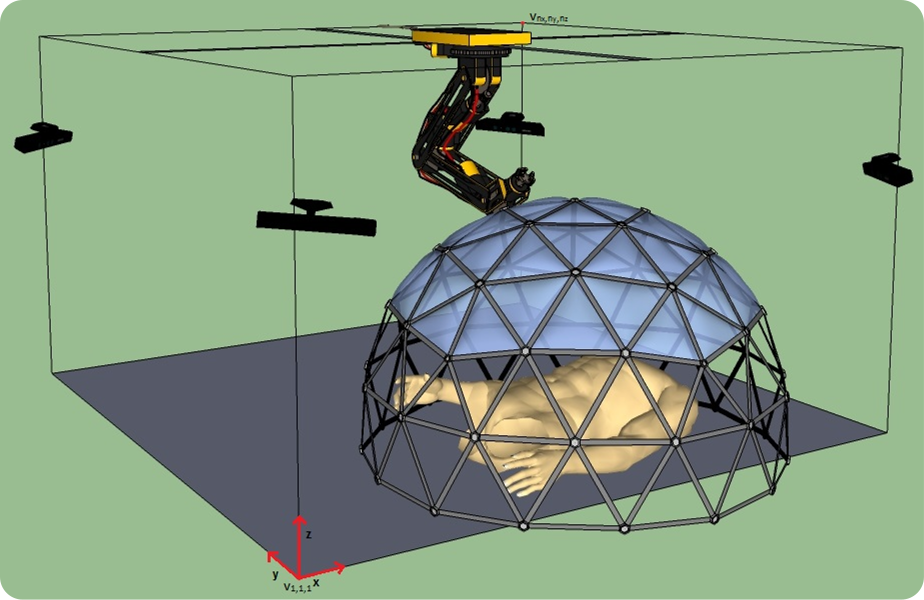}
  \centering
  \caption{Depth sensor setup for tabletop setting.}
  \label{fig:setup}
\end{figure}

The key to our algorithm for Dynamic Next Best view (DNBV) is reconstructing the dynamic 3D scene with data captured from sensors that observe it as in Fig.~\ref{fig:setup}. 
Here, four Kinect V2 sensors are used to capture activity in a workcell setting. 
The voxel map described in the Sec.~\ref{sec:occmap} is used for this purpose. 
Then a constrained tessellated view sphere is initialized similar to the ones suggested in~\cite{Massios1998, Morooka1998}, presented in Sec.~\ref{sec:tess}.
Next the voxel map is projected onto the tessellated sphere and faces that correspond to sets of unoccluded projected voxels are identified as candidate viewpoints. 
Then a new position for the robot arm can be predicted by maximizing an objective function that will take into account the occlusion, the viewpoint quality, and finally how far the sensor will have to move. 

In our case there is a camera mounted on a robot arm. Similar ideas to the ones described above could also be found in \cite{Hunt1982,Papanikolopoulos1993}. 
The above procedure is summarized in Fig.~\ref{fig:scheme}.

% #######################################################
\section{Initialization}\label{sec:inits}
This section will describe how to initialize the tessellated hemisphere and the occupancy grid.
It also explains the indexing procedure that is used for the occupancy grid.

%-----------
\subsection{Tessellated view sphere}\label{sec:tess}

\begin{figure}[b]
  \includegraphics[width=\columnwidth]{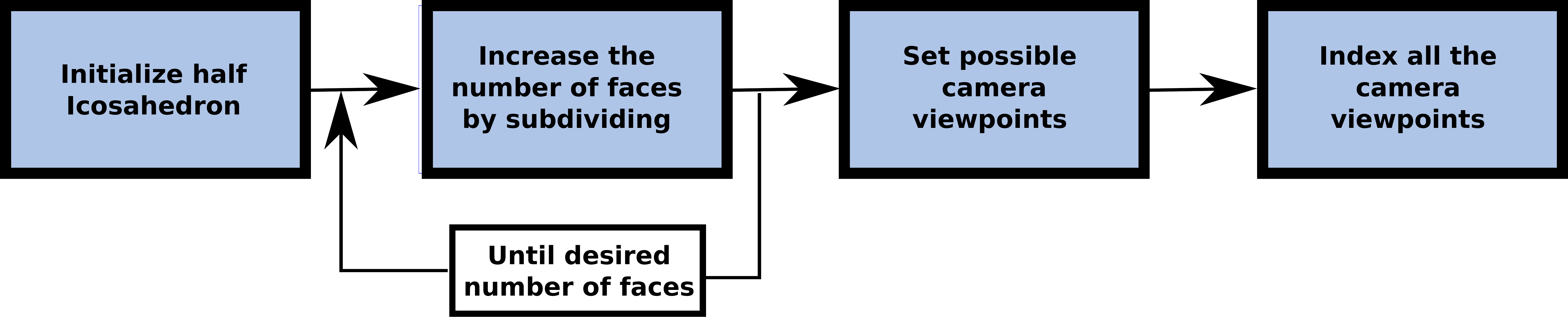}
  \centering
  \caption{Scheme of tessellated hemisphere initialization.}
  \label{fig:tess-init}
\end{figure}

As one task-based constraint is to avoid interfering with the people in the workcell, the camera is assumed to be at some distance from the target. 
Here, the surface of a sufficiently large hemisphere defines the 2 DOF space of camera positions (azimuth, elevation), assuming the camera always faces upward.

In general the space of possible views is continuous, but here we limit and discretize the space to the surface of a tessellated sphere, also referred as the dome in this paper. 
We restrict the movement of the robot arm such that camera mounted on the end effector moves on the surface. 
To obtain a discrete set of possible positions we tessellate the sphere, such that all the positions are equally likely to be chosen. 
In the literature there are many ways to tessellate a sphere. 

Here, the set of potential viewing positions is defined by the face centers of a half icosahedron centered above the target in the workcell. 
The highest point of the dome is the north pole and the set of the lowest points is the equator. 
The facets lying on the equator are removed, since we will restrict attention to the upper area of the dome as described in the next section which presents the objective function. 

The center of each facet is a possible camera position. 
With the initial dome there are only 10 possible camera positions. 
In order to generate more viewpoints, each face is subdivided into four equal pieces, by connecting the centers of the edges of each triangle. 
This is repeated until the desired viewpoint resolution is achieved. 
Indexing each camera position starts from one of the closest positions to the north pole by labeling it as $cp_{1}$ then in a spiral passing through all the unlabeled points until the equator is reached. 
The generation of the tessellated hemisphere is summarized in Fig.~\ref{fig:tess-init}.
% \begin{enumerate}
%   \item Initialize an icosahedron.
%   \item Split into two equal parts and keep the top part.
%   \item Increase the number of faces by subdividing triangles.
%   \item Repeat 3 until you reach the desired number of faces
%   \item Set the centers of each face as a possible camera viewpoints.
%   \item Index all the camera viewpoints.
% \end{enumerate}

\begin{figure}[b]
  \includegraphics[width=\columnwidth]{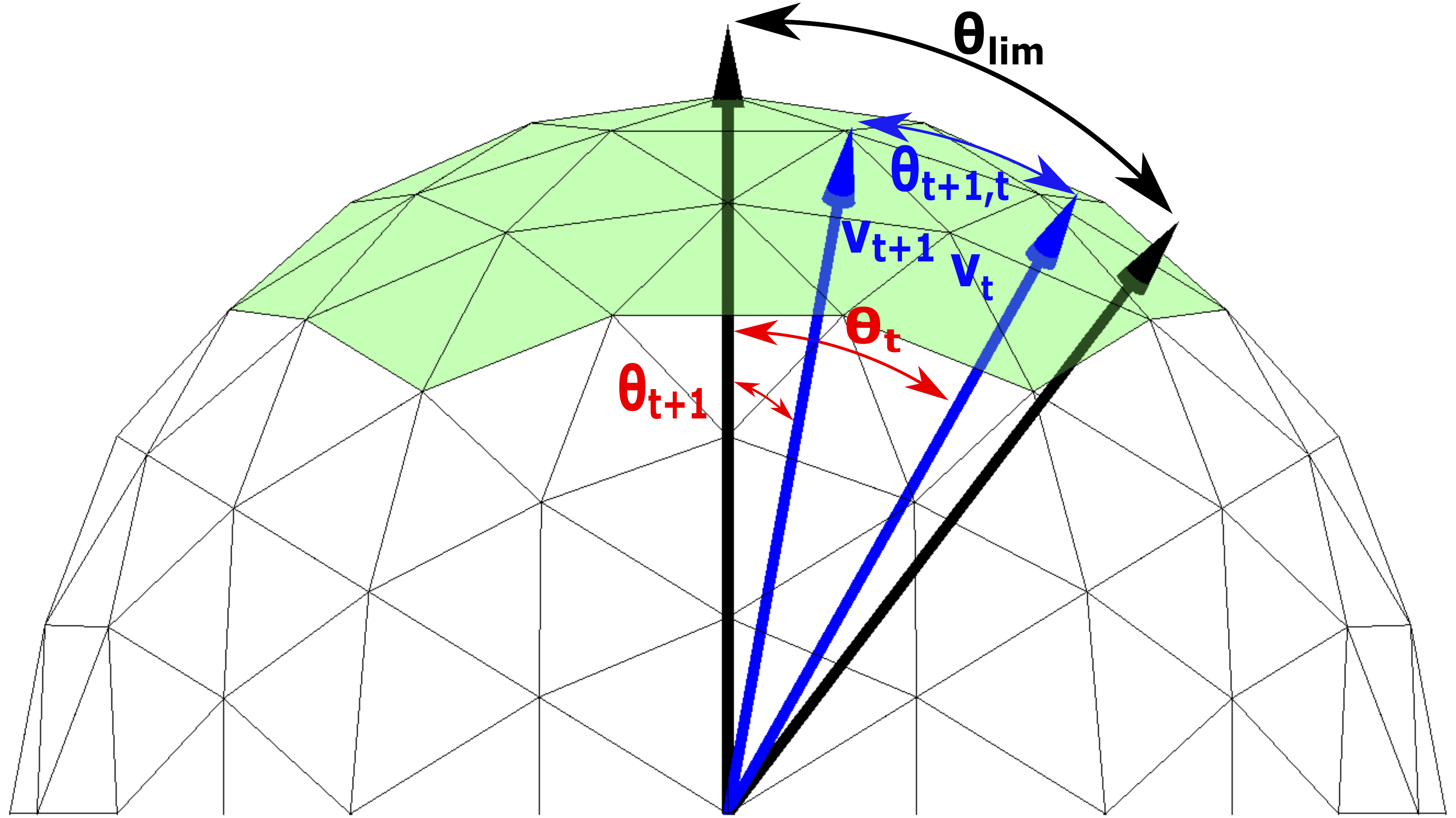}
  \centering
  \caption{Tessellated hemisphere (dome) with two camera poses for the current ($t$) and next ($t+1$) state. 
  The $\theta_{lim}$ is the maximum allowed angle, $\theta_{t}$ and $\theta_{t+1}$ are the angles that camera viewpoints at states $t$, $t+1$ make with the normal to the table and $\theta_{t+1,t}$ is the angle between viewpoints $v_{t}$ and $v_{t+1}$.}
  \label{fig:dome}
\end{figure}

%-----------
\subsection{Voxel map}\label{sec:occmap}
In the literature a common tool used in the implementation of NBV to identify occluded voxels is a voxel map. 
According to \cite{Massios1998} the voxel map is a volumetric representation of the scanned space including the object we are interested in. 
Each voxel is a fixed volume cube whose size is pre-specified. 
The volume of each voxel is based on the width, height and depth of the scene. 
Then it is discretized to $n_{x}$, $n_{y}$ and $n_{z}$ points for $x$, $y$ and $z$ dimension respectively. 
The $v_{xyz}$ identifies each voxel, then the occupancy grid is denoted as 
$X_{t}=\{v_{xyz}\}$, where $x=1,\ldots, n_{x}$, $y=1,\ldots, n_{y}$ and $z=1,\ldots, n_{z}$.

Apart from identifying each voxel we also need a systematic labeling when assigning the points of the point-cloud to each voxel. 
Sanchiz et al.~\cite{Sanchiz1999} describe a straightforward method to achieve that using the round-scale function. 
The labeling of each voxel varies from one author to the other but all are similar. 
Those labels could be discrete (ie. only one possible label) or probabilistic~\cite{Hornung2013}.
Here, the following discrete set of labels for $v_{xyz}$ is used:
\begin{itemize}
  \item \textit{Empty}. These voxels are empty and are normally between the sensor and the surface of the scanned object
  \item \textit{Seen}. These are normally the voxels that include the surface of the scanned scene, including any occluding objects. For the voxels with this labeling we also estimate quantities related to region quality as well as the surface normal.
  \item \textit{Unseen}. The voxels that have not been scanned.
\end{itemize}
The seen voxels of $X_t$, represented with their center point, are then projected onto the tessellated sphere to determine target visibility and viewpoint occlusion. 
To do that, the tessellated dome is approximated with a hemisphere whose origin lies just above the target point (and so here contains only occluding voxels). 
For each voxel center $v_{xyz}$ inside the dome the polar and azimuth angles $\phi$ and $\theta$ of its projection onto the hemisphere are computed. 
Hence given a radius its coordinates on the hemisphere can be calculated. 
Then for each projected point the geodesic distance to all dome vertices is computed and the three closest are found, which allows identification of the face $i$ that the point projects to\footnote{There is probably a more efficient algorithm than computing all distances.}. 
In total $m^i_t$ denotes the number of such 3D points that project to face $i$.
How these projected points are used is discussed in Sec.~\ref{sec:target-occ}.

\subsection{Instantiation of the occupancy grid}

There are a number of methods for instantiating the values of an occupancy grid~\cite{Hornung2013}.
Irrespective of the sensor used, it is assumed that the data is a 3D point cloud $\{\vec{x}_i\}$ representing the scene.
The occupancy grid is instantiated as \textit{unseen} and then all voxels containing a 3D point are marked as \textit{seen}.
Finally, all voxels lying on the ray from the viewpoint center to the surface past the final \textit{seen} voxel are marked as \textit{empty}.

Note that as this is a dynamic problem. The occupancy grid is recalculated each time a new point cloud is available.
This differs from the standard NBV algorithm which fuses the new point clouds instead of replacing them.

% #######################################################
\section{Objective function}\label{sec:obj-fun}
This section introduces the objective function used to predict the next best view. 
The possible camera locations (Fig.~\ref{fig:dome}) are described with a set of parameters: 
$r$ is the radius of the dome, 
$\vec{n}$ is the normal of the workcell surface, 
$\vec{v}_{b}$ the direction of the view of the camera when on the surface of the dome at point $b$ and 
$a$ is the maximum allowed viewpoint angle.

To model the problem the current state $t$ is described with the following parameters: 
\begin{equation}
s_{t} = \{\vec{v}_{t}, \vec{j}_{t}, \vec{m}_{t}\},
\end{equation}
where $\vec{v}_{t}$ is the direction of the viewpoint (unit vector),  
$\vec{j}_{t}$ are the joint parameters of the robot arm and the 
$\vec{m}_t = \left\lbrace m^i_t \right\rbrace $ is a vector that counts the voxels occluding each each view. 

The objective function is a linear combination of probability-like distributions that will account for visibility, occlusion level and distance that needs to be traveled by the camera in joint and 3D space. 
The algorithm optimizes the choice of a position where the viewing angle is acceptable, the number of occluded voxels is small and which is as close as possible to the previous camera position.
These distributions over the viewpoints on the viewsphere sum to 1. They are used to normalize the different components of the objective function. 

The goal is to choose parameters of a state $s_{t+1}$ that maximize the following objective function:
\begin{equation}
\begin{split}
P_{total}(s_{t+1}\mid s_{t}) = w_{vis}P_{vis}(s_{t+1}) + w_{nocc}P_{nocc}(s_{t+1}) + \\
+ w_{dist}P_{dist}(s_{t+1}\mid s_{t}) + w_{jt}P_{jt}(s_{t+1}\mid s_{t})
\end{split}
\label{eq:P-total}
\end{equation}
where $w_{vis}$, $w_{nocc}$, $w_{dist}$ and $w_{jt}$ are the weights controlling the mixture of the distributions 
$P_{vis}(s_{t})$, $P_{nocc}(s_{t+1})$, $P_{dist}(s_{t+1}|s_{t})$, and $P_{jt}(s_{t+1}|s_{t})$ 
that are responsible for visibility, occlusion and distance to travel in the 3D and joint spaces respectively. 

The transition between the states is described with the variables of $s_{t}$ and $s_{t+1}$. 
In the distributions we will include only variables on which they actually depend on to avoid notation overload.

The distribution that controls the angular distance traveled on the dome is based on the variable 
\begin{equation}
  \Delta \theta_{d} = \cos^{-1}(\vec{v}_{t} \cdot \vec{v}_{t+1}),
\end{equation}
which corresponds to the angle between the viewpoints at states $t$ and $t+1$. 

The remaining part of this section presents the probability distributions that will account for the factors we are interested to control.

%-----------
\subsection{Viewpoint quality}
The best overview of the scene is provided by the top viewpoint, which brings us to discourage solutions close to the boundary of the restricted dome (area within $\theta_{lim}$ view angle).
All possible positions in the state $t+1$ that lie inside are given by the viewpoint angle~$\theta_{t+1}$ in
\begin{equation}
  \theta_{t+1}=\cos^{-1}(\vec{n} \cdot \vec{v}_{t+1}) \leq \theta_{lim},
\end{equation}
where the limit angle $\theta_{lim}$ between $\vec{n}$ and $\vec{v}_{b}$ is given by 
\begin{equation}
  \theta_{lim} = \cos^{-1}(\vec{n} \cdot \vec{v}_{b}).
\end{equation}
It is used to scale the variables that control distance and angle. 
Thus the following distribution is used:
\begin{equation}
  P_{vis}(s_{t+1}) = P_{vis}(\theta_{t+1} ) = 
    \frac{1}{Z_{vis}} e^{ -\frac{1}{2} \big(\frac{2\theta_{t+1}}{\theta_{lim}} \big)^2}\,,
\end{equation}
where $ 0\,\leq\, \theta_{t+1} \,\leq\, \theta_{lim}$ and $Z_{vis}$ is a normalization factor for the distribution. 
This formula favors angles less than $\frac{1}{2}\theta_{lim}$.  

%----------
\subsection{Target occlusion} \label{sec:target-occ}
Recall that $\vec{m}_{t+1}$ is the vector that contains number of projected occluding voxels for each camera position $i$ as described in Sec.~\ref{sec:occmap}.
The $P_{nocc}(s_{t+1})$ models whether a viewpoint has too much occlusion or not. 
If the number of occlusions seen from the camera position exceeds a specified threshold, the viewpoint is classified as occluded and $P_{nocc}(s_{t+1})=0$ for that particular viewpoint, otherwise $P_{nocc}(s_{t+1}) = 1$ marks a good viewpoint.
Then 
\begin{equation}  
  P_{nocc}(s_{t+1}) =  
     \begin{cases}
       1 & \text{if} \,\,m_{t+1}^i < m_{0},\\
       0 & \text{otherwise},\\
     \end{cases}
\end{equation}
where $m_{0}$ is a threshold in the number of occluding voxels above which the viewpoint is considered occluded.
%----------
\subsection{Travel distance} \label{sec:distance}
It is desirable to minimize the travel distance between two points that lie on the sphere so that the view does not change much. A distribution that will help us to minimize the distance between the current and predicted viewpoint is given below. 
The distance $d_{t+1,t}$ between points at states $t$ and $t+1$ that lie on a sphere is given by 
\begin{equation}
  d_{t+1,t}=r\theta_{t+1,t},
\end{equation}
where $r$ is the radius of the dome and $\theta_{t+1,t}$ can be computed by
\begin{equation}
  \theta_{t+1,t} = \cos^{-1}(\vec{v}_{t+1} \cdot \vec{v}_{t}).
\end{equation}
Using similar reasoning to the visibility distribution the distance distribution is:
\begin{equation}
 P_{dist}(s_{t+1}|s_{t})=P_{dist}(d_{t+1,t}) = \frac{1}{Z_{dist}} e^{-\frac{1}{2} \big(\frac{d_{t+1,t}}{f(\theta_{lim})} \big)^2},
\end{equation}
where $0 \,\leq \,d_{t+1,t} \,\leq\, 2\theta_{lim}r$. $f(\theta_{lim})$ and $Z_{dist}$ are scaling factors for the random variable $d_{t+1,t}$ and the distribution respectively. 
In particular we set $f(\theta_{lim})=\frac{\theta_{lim}}{2}r$. 
Since the radius of the dome is fixed $P_{dist}$ can be simplified:
\begin{equation}
  P_{dist}(d_{t+1,t}) = 
  \frac{1}{Z_{dist}} e^{-\frac{1}{2} \big(\frac{2 \theta_{t+1,t}}{\theta_{lim}} \big)^2} \,,
\end{equation}
where $0 \,\leq \,\theta_{t+1,t} \,\leq\, 2\theta_{lim}$.  
The reason for scaling the random variable $\theta_{d}$ with $\frac{1}{2}\theta_{lim}$ is similar as before. 
In this case though, the scaling results in a much narrower distribution to favor smaller motions and thus better visual stabilization.

\subsection{Robot arm joint parameters}
As well as minimizing the sensor motion (see Sec.~\ref{sec:distance}) it is also desirable to minimize the robot joint angle changes, so as to limit large changes in joint space configuration. 
This section gives the distribution for small transitions in the joint space between the set of joint parameters for the current and the predicted positions. 

Let $\mathbb{J}\subseteq \mathbb{R}^{n}$ be the set of all allowed joint parameters that control the robot arm. 
For the robot arm we are using there are $n=6$ parameters. 
Let $\vec{j}_{t}, \vec{j}_{t+1} \in \mathbb{J}$ be the joint parameters responsible for the current and predicted states of the robot arm respectively. 
Following a similar reasoning to the distance and angle the distribution $P_{jt}(s_{t+1}|s_{t})$ is defined as:
\begin{equation}
\begin{split}
  P_{jt}\bigl(\,\vec{j}_{t+1} \,\, | \,\,\vec{j}_{t},\Sigma \bigr) =
  \frac{1}{Z_{jt}} e^{-\frac{1}{2} (\vec{j}_{t+1}-\vec{j}_{t})^{\intercal}\Sigma^{-1}(\vec{j}_{t+1}-\vec{j}_{t})},
\end{split}
\end{equation}
where $Z_{jt}$ is a normalization constant for the probability distribution. 
Furthermore, $\Sigma$ is a diagonal approximation of the covariance matrix of the form $\Sigma=\sigma_{jt}^2 \mathbb{I}$,
where $\sigma_{jt} ^2$ is the variance of the distances in joint space between all camera positions.

% ##########################################################
\section{Optimization}

This section explains how the parameters of the objective function~\eqref{eq:P-total} are determined. In the first phase the cost weights $\vec{w}$ are trained to describe the desired behavior. 
This fully specifies the objective function, which can be in turn optimized to predict the next state at run-time.

\subsection{Estimation of cost weights}
To avoid having to manually assign ground truth predictions for a large number of viewpoints to train the cost weights, we instead prescribe the value of the cost function, which allows to describe the desired behavior more flexibly. 

We use cross validation to train the weights $\vec{w}$ by minimizing a sum of squares distance between $P_{total}(s_{t+1}|s_{t})$ and the prescribed function $\tilde{P}_{t+1,t}$ for all allowed states $t$ and $t+1$. 
The score for each new state is given by the scoring function
\begin{equation}\label{eq:target-cost}
\tilde{P}_{t+1,t} = \alpha_{s} \Omega_{t+1} + \alpha_{d} e^{-d_{t+1,t}} + 
  \alpha_{\theta}e^{-\theta_{t}},
\end{equation}
where $\Omega_{t+1}$ evaluates to 1 when the viewpoint associated with $s_{t+1}$ is not occluded (0 otherwise), the $d_{t+1,t}$ is the distance moved between the viewpoints and $\theta_t$ encourages a more vertical viewpoint.

Weights $\alpha_{s}$, $\alpha_{d}$ and $\alpha_{\theta}$ control the behavior of the objective function giving priority to either avoiding occlusion $\alpha_{s}$, moving to closer viewpoints~$\alpha_{d}$ or higher viewpoints~$\alpha_{\theta}$. 
We use $e^{-x}$ for both angle and distance because it takes larger values for smaller values of $x$ and it drops exponentially for larger ones. 
Hence when the weights of $P_{total}$ are trained with score $\tilde{P}_{t+1,t}$ it will pick neighboring viewpoints that will meet different criteria depending on the combination of the weights. 
In the experimental section, we explain how we choose the most appropriate set of weights.

To select the $\alpha$ weights of the objective function, we set a desired behavior which is controlled by the weights~$w$ of~(\ref{eq:target-cost}). 
To explore different behaviors of $P_{total}$ we initialize $[\alpha_{s}, \alpha_{d}, \alpha_{\theta}]$ with values from the set $\{0.5,1,1.5,2\}$ in all possible combinations allowing repetition. 
Hence there are $64$ possible sets of weights and we look for the values that best produce the smooth behavior we seek. 
To choose the most appropriate behavior we monitor the number of jumps the algorithm makes, the average distance traveled and the average increase in the $z$ direction (we want to move upward for better viewpoints) when~$s_{t}$ differs from~$s_{t+1}$. 

Given the $\alpha$ values, we then find the weights of the objective function  (\ref{eq:P-total}) by picking the set of weights $\vec{w}=[w_{nocc},w_{vis},w_{dist},w_{jt}]$ that approximates the desired behavior. 
These weights $\vec{w}^{\,\ast}$ are those that optimize (\ref{eq:weight-train}) on both training and testing sets:
\begin{equation} \label{eq:weight-train}
  \vec{w}^{\,\ast} =  
     \arg \min_{\vec{w}}\sum_{i,j \in S} 
     \big(P_{total}(s_i \mid s_j)-\tilde{P}_{i,j} \big)^2,
\end{equation}
where $S$ is the set of considered viewpoints and $i,j$ are all pairs of viewpoints corresponding to states $t+1$ and $t$ respectively. 
To find $\vec{w}$ we initialized a set of weights ranging from $0$ to $1$ from a uniform distribution and normalize their sum to 1. 
Then we find the minimizer of (\ref{eq:weight-train}) using Matlab's built in function \texttt{fmincon}.  
%To find $\vec{w}$ we sample 5000 sets of weights ranging from $0$ to $1$ from a uniform distribution and normalize them to 1. Then we look for the minimizer of  (\ref{eq:weight-train}).

We have observed that a good performance is achieved when $w_{vis} \ll w_{dist}$, which is the case of the values actually used in the experiments (Sec.~\ref{sec:exp-params}).

%From the data we can observe a correlation between the weights of the scoring and objective functions. There are two clear patterns that emerge from the data. 
%In the case where $w_{vis} \ll w_{dist}$ the algorithm decides moving to a neighboring viewpoint only when the current viewpoint becomes occluded. 
%This may result in the algorithm getting trapped in local maxima or on the boundary until the occlusion status of the viewpoint changes. 
%In the case where $w_{dist} \ll w_{vis}$ the algorithm decides to move to neighboring viewpoints until it reaches the upper positions of the tessellated dome (i.e. better viewpoints). 
%The final case is when $w_{dist} \sim w_{vis}$ and does not differ much from the previous ones. 
%Depending on the larger weight the algorithm will choose either to move to a nearby position in case the current viewpoint becomes occluded or to jump to a better viewpoint. 
%The difference is that $w_{jt}$ becomes more significant. 
%Hence the transitions to the viewpoint (influenced by the angle and distance contributions in the objective function) take place if the viewpoints are not far in the joint space provided that both states are not occluded. 

\subsection{Inference of the next state}
With the weights cost determined as in the previous section we can optimize if and where the camera should move given the current state. 
The next state $s^{\ast}_{t+1}$ is chosen to maximize the function \eqref{eq:P-total} in
\begin{equation}
  s^{\ast}_{t+1} = \arg\max_{s_{t+1}} P_{total}(s_{t+1}\mid s_{t}). 
  \label{eq:optimization}
\end{equation}
In our case the discrete set of positions can be simply enumerated to find the state with maximum value.

% #######################################################
\section{Experiments}\label{sec:exp}
In this section we present the results for the sets of weights both for the scoring and objective function. 
We discuss the different effects that each set of weights for the scoring function produces and choose the most appropriate ones to train $P_{total}$. 
To fuse the data from multiple depth sensors we use OctoMap~\cite{Hornung2013}, which filters the noise occurring due to temporal instability, interference or calibration inaccuracies, and produces an octree voxel map. 

\subsection{Experimental setup}
The following apparatus depicted in Fig.~\ref{fig:setup} is used:
\begin{itemize}
  \item workcell with table and supporting structure,
  \item four Kinect V2 depth sensors facing the region of interest, calibrated extrinsically using a pattern,
  \item UR 10 robot arm with 6 DOF top-mounted above the center of the table, 
  \item camera mounted on the end-point of the robot arm,
  \item controlling computer with i7-6700 @ 3.4 GHz CPU and 16Gb of RAM.
\end{itemize}

Position $v_{1,1,1}$ corresponds to the bottom left corner wrt. the observer and  $v_{n_{x},n_{y},n_{z}}$ to the opposite top corner of the workcell space.  
 We initialize the dome of radius $0.7$ meters with two subdivisions of the initial icosahedron and obtain $n_1=44$ allowed camera poses in the upper part of the dome 
(see Fig.~\ref{fig:dome-scenes} ).

The experiments cover 8 different scenarios, 6 acquired with the setup described above and 2 simulated cases. 
For the simulated cases we manually occluded camera viewpoints to create specific extreme situations.
For training the objective function weights we use cross validation with a training set $50\%$ of the eight scenarios, ie. we trained on ${{8}\choose{4}}=70$ cross-validation splits. 

To decide whether a viewpoint is occluded we experimentally set the occupancy threshold to $m_{0}=3$. 

%--------------
\subsection{Experimental parameters}\label{sec:exp-params}

Since our goal is to record a sequence, we not only need the best viewpoint, but also to have reasonably stabilized images. 
In this setup we focus on the case $w_{vis} \ll w_{dist}$ because it fits our requirements. 
To have the smoothest possible video recording and joint parameter transitions we choose the following set of weights for the score
\begin{equation}\label{eq:final-weights}
[\alpha_{s},\alpha_{d}, \alpha_{\theta}]=\big[ 0.5, 1, 2 \big]
\end{equation}
that minimizes the number of jumps and the average distance traveled, and jump in $z$ direction for consecutive states. 
Optimizing  (\ref{eq:weight-train}):
\begin{equation}
[w_{nocc},w_{vis},w_{dist},w_{jt}]=[0.080, 0.238, 0.585, 0.096]
\end{equation}
 
Using this set of weights results in average geodesic distance of $14.4$ cm.
The training took approximately $2.5$ minutes using a single processor. 
To calculate NBV the algorithm requires 50 ms in the Matlab implementation. 

\subsection{Evaluation}
\begin{figure}
  \centering
  \includegraphics[width=0.8\linewidth]{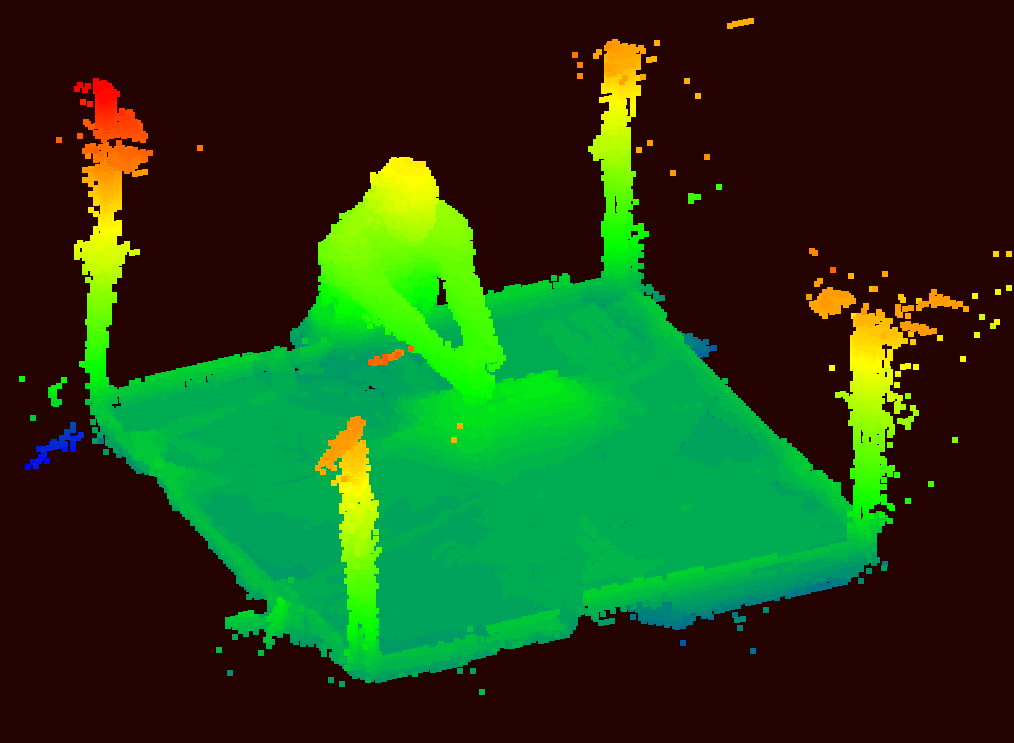}
  \caption{Example fused point cloud captured by the depth sensors, height colored.}
  \label{fig:point-cloud}
\end{figure}

\begin{figure} 

%  \begin{subfigure}[b]{\columnwidth}
%    \begin{center}
%        \includegraphics[width=0.48\linewidth]{tess_dome_2.pdf}
%        \caption{Tessellated dome}
%        \label{fig:dome-scenes:orig}
%    \end{center}
%  \end{subfigure}
   
  \begin{subfigure}[b]{0.48\columnwidth}
    \begin{center}
        \includegraphics[width=\linewidth]{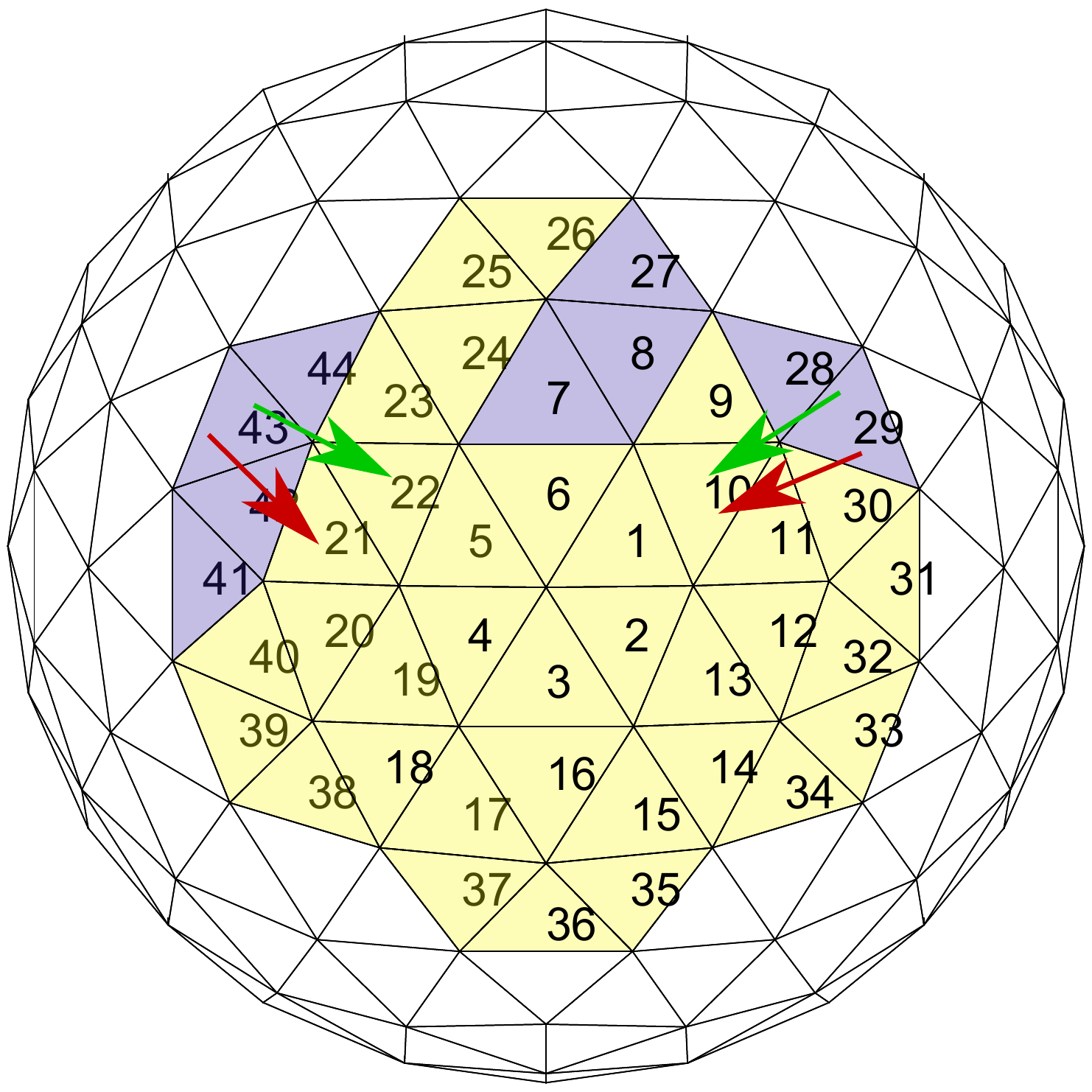}
		\caption{scene 4}
        \label{fig:dome-scenes:scn4}
    \end{center}
 \end{subfigure}
 \begin{subfigure}[b]{0.48\columnwidth}
   \begin{center}
        \includegraphics[width=\linewidth]{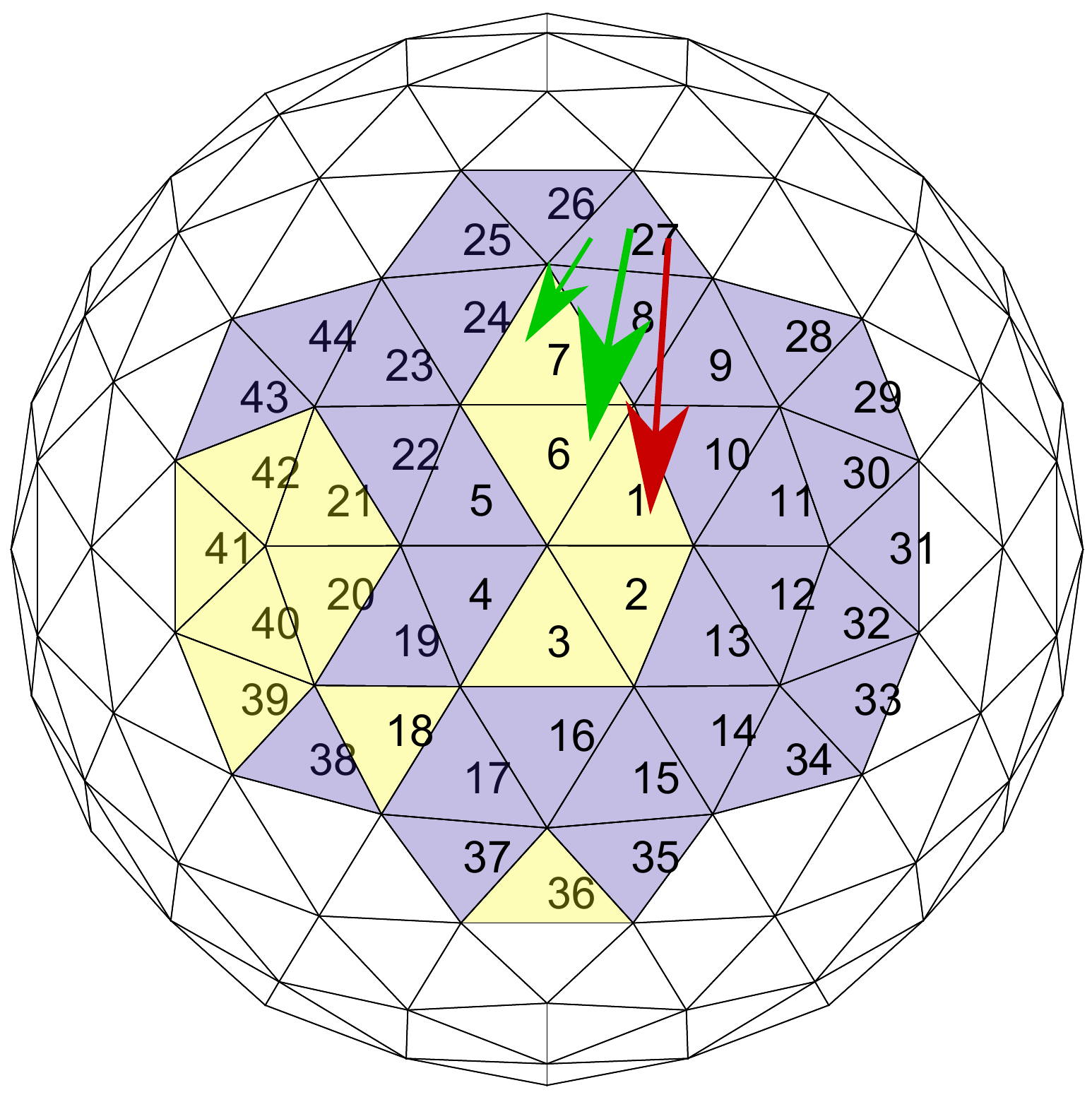}
        \caption{scene 7}
        \label{fig:dome-scenes:scn7}
    \end{center}
  \end{subfigure}

  \caption{Enumeration of the 44 viewpoints on the dome. The numbers are only on the allowed camera viewpoints. Color indicates occlusion of viewpoints (purple: occluded, yellow: not occluded). Green and red arrows indicate the new position according to GT and algorithm prediction respectively (see text). }
  \label{fig:dome-scenes}
\end{figure}

We have 8 different scenes represented by point clouds with different levels of occlusion.
An example is given in Fig.~\ref{fig:point-cloud}. 
In each one we compared the output of our algorithm against the ground truth. 
The ground truth has been manually assigned, i.e. for each case a human supervisor has labeled the next best viewpoint. 
The limitation of the established ground truth is that the person cannot take into account joint space constraints.   
For each viewpoint in each scenario we estimate the best next view point regarding the occlusion, viewpoint, joint and geodesic distance. 
For each scenario there are $44$ possible initial viewpoints (see Fig.~\ref{fig:dome-scenes}) of which there are three non-reachable viewpoints (6,7,8) due to robot arm limits. 
We test the behavior of the algorithm for each case, as if every viewpoint would be the current state.
The dataset with total $41\times 8 = 328$ cases (of initial and next view) can be downloaded from the project website\footnote{\url{http://homepages.inf.ed.ac.uk/rtylecek/DNBV}} .

There are 16 cases out of the 328 where the algorithm provided different output than the ground truth. 
For these 16 viewpoints we found out that either the position in the ground truth could not be reached because of the joint limits, or the joint configuration drastically changes because the arm has to flip over, i.e. large joint distance. 

One example is given for Scn.~4 viewpoints 43 and 29. 
They are symmetric and occluded, as shown in Fig.~\ref{fig:dome-scenes:scn4}. 
The ground truth expected them to move to viewpoints 22 and 10 respectively (green arrow). 
However, our algorithm predicted the move from 43 to 21 instead (red arrow), due to the joint distance, while viewpoint 29 moves to 10 as expected. 
Another interesting case is in Scn.~7 (Fig.~\ref{fig:dome-scenes:scn7}). 
If we initialize from point 27, the algorithm results in viewpoint 1 instead of 6 or 7. 
This situation occurs because viewpoint 6 and 7 cannot be reached due to joint limits.

Using Table \ref{tab:error-analysis} we can justify the outputs of the algorithm. In the case where the algorithm decides to jump from viewpoint 43 to 21 it is clear that 21 beats 22 in all criteria but angle. But we consider the objective function weights that give higher priority to distance to have a smoother video. When the camera is initialized in viewpoint 27 in point cloud 7, viewpoints 6 and 7 are selected, but they are not reachable because of robot joint limits (see Fig.~\ref{fig:dome-scenes:scn7}). 
\begin{table}
\begin{center}
\begin{tabular}{|c|cc|ccc|}
\hline
Scn. & Start & End & Distance & Joints & Angle \\ \hline \hline
4      & 43        & 21       & 0.30  & 0.17     & 0.44  \\
4      & 43        & 22       & 0.36  & 0.24     & 0.35  \\
7      & 27        & 1        & 0.54  & 1.87     & 0.17  \\
7      & 27        & 6        & 0.48  & -      & 0.17  \\
7      & 27        & 7        & 0.31  & -      & 0.36  \\ \hline
\end{tabular}
\end{center}
\caption{Analysis of points that do not match the ground truth in the two example scenarios (Scn. 4 and 7).
The columns above show the initial viewpoint, the final viewpoint, the geodesic and joint distance traveled from initial to final view point and viewpoint angle respectively.
The `-' denotes points that are not reachable due to robot joint limits.}
\label{tab:error-analysis}
\end{table}

\begin{figure*}[!ht]
\begin{center}
\begin{subfigure}[b]{.3\textwidth} \centering
\includegraphics[width=0.75\textwidth]{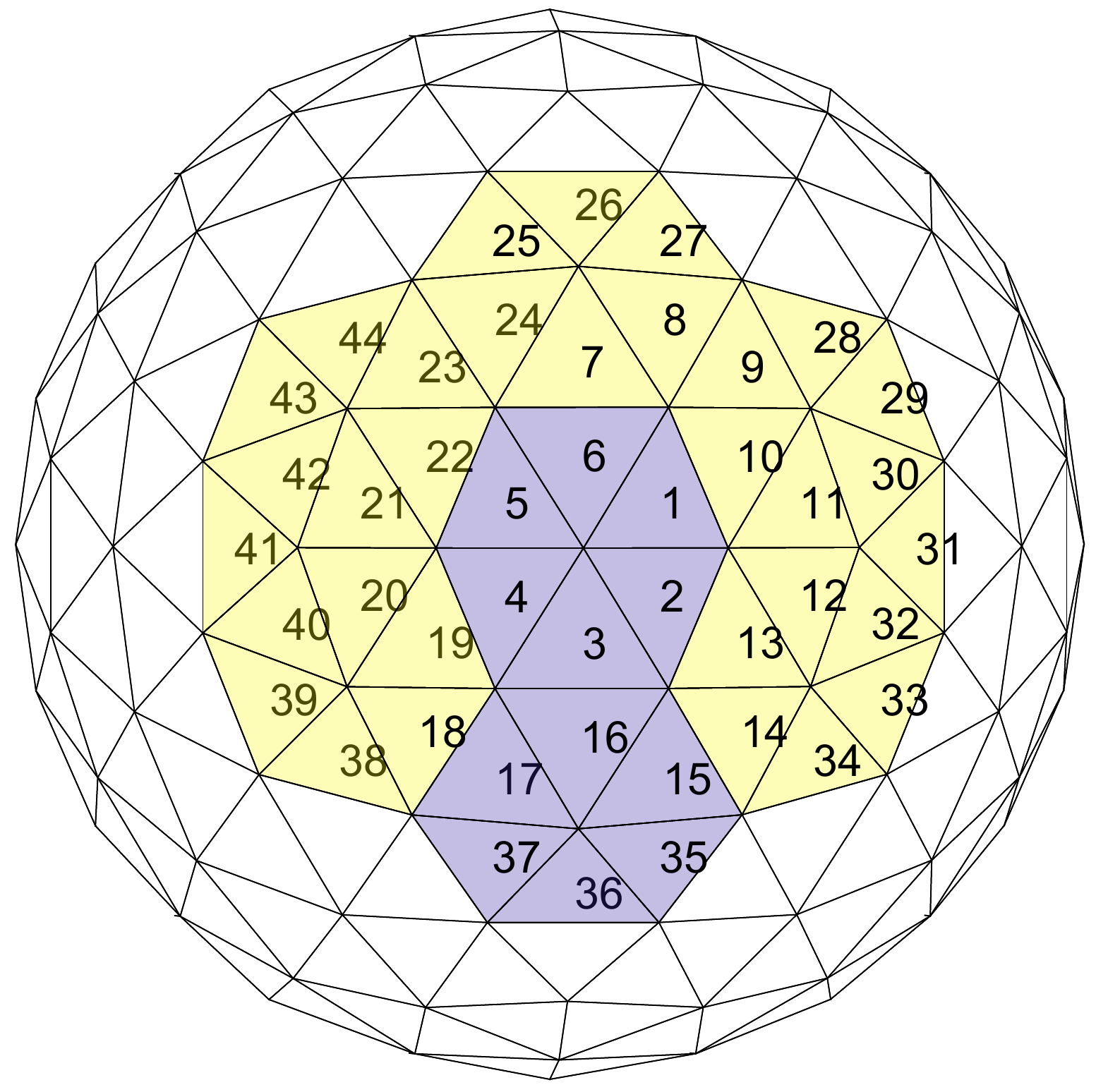}
        \caption{top occlusion}
        \label{fig:dynamic1}
\end{subfigure}
\begin{subfigure}[b]{.3\textwidth} \centering
\includegraphics[width=0.75\textwidth]{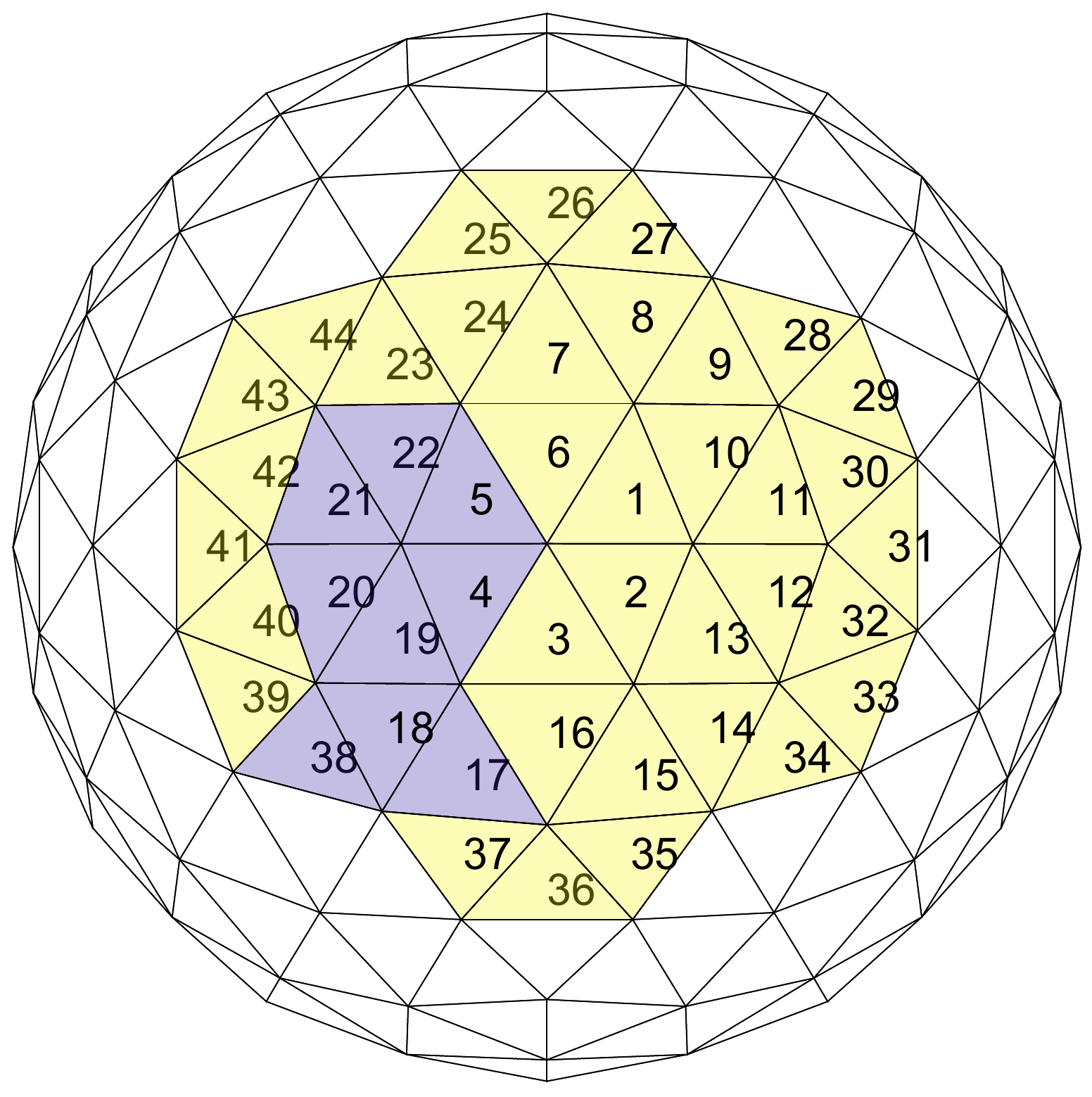}
        \caption{left occlusion}
        \label{fig:dynamic2}
\end{subfigure}
\begin{subfigure}[b]{.3\textwidth} \centering
\includegraphics[width=0.75\textwidth]{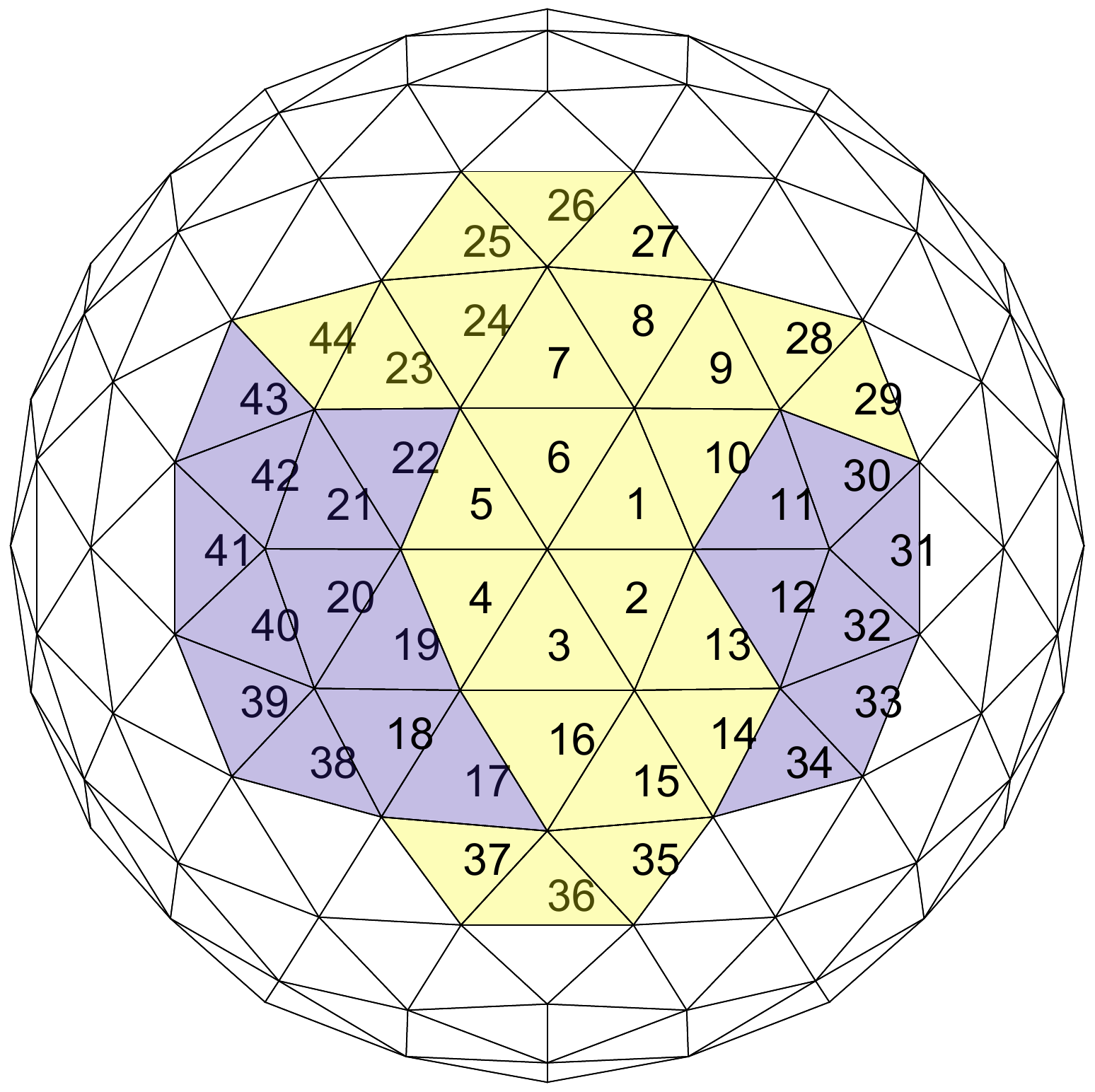}
        \caption{both sides occluded}
        \label{fig:dynamic3}
\end{subfigure}
\begin{subfigure}[b]{.3\textwidth} \centering
\includegraphics[width=0.75\textwidth]{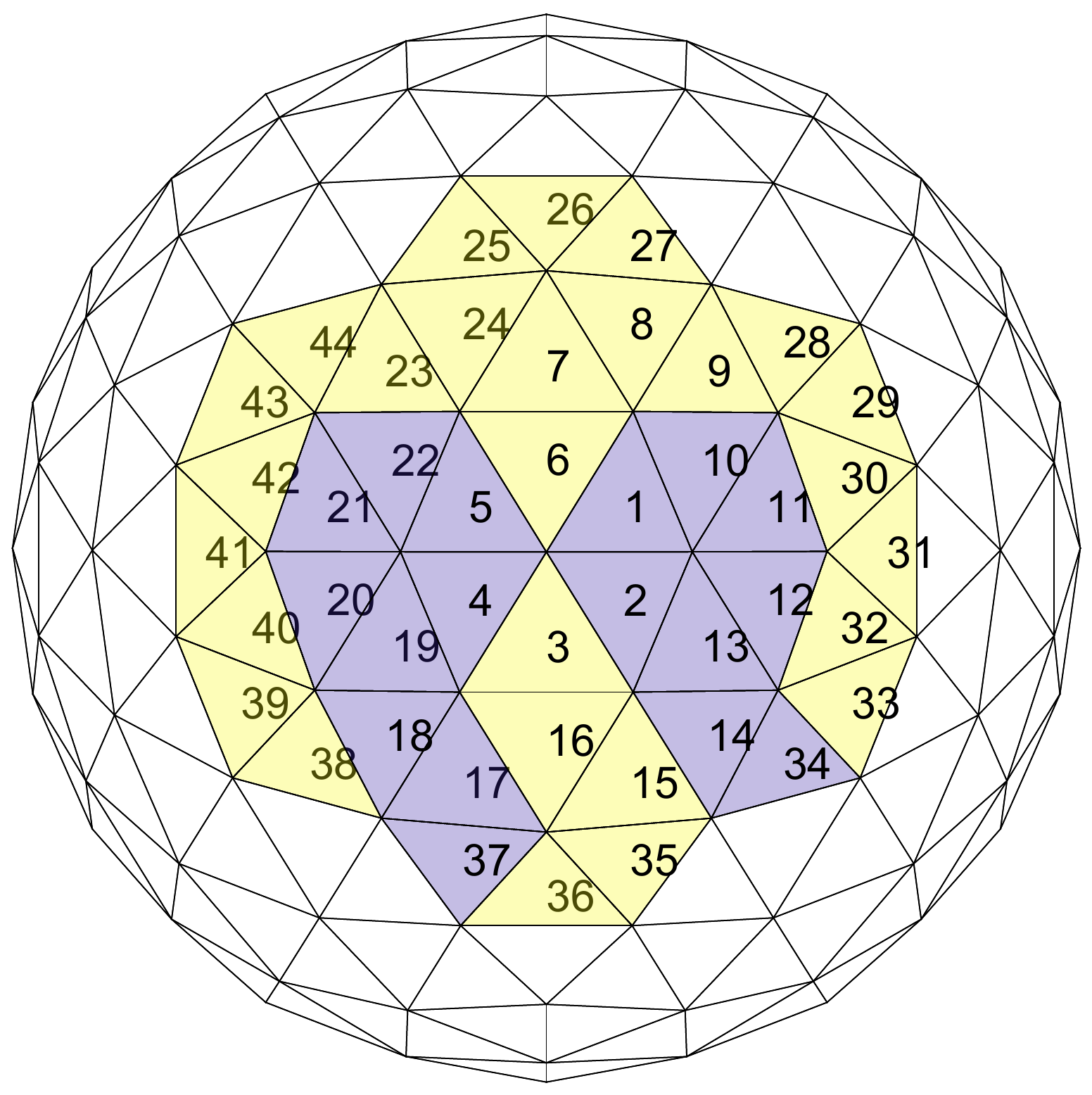}
        \caption{two hands}
        \label{fig:dynamic4}
\end{subfigure}
\begin{subfigure}[b]{.3\textwidth} \centering
\includegraphics[width=0.75\textwidth]{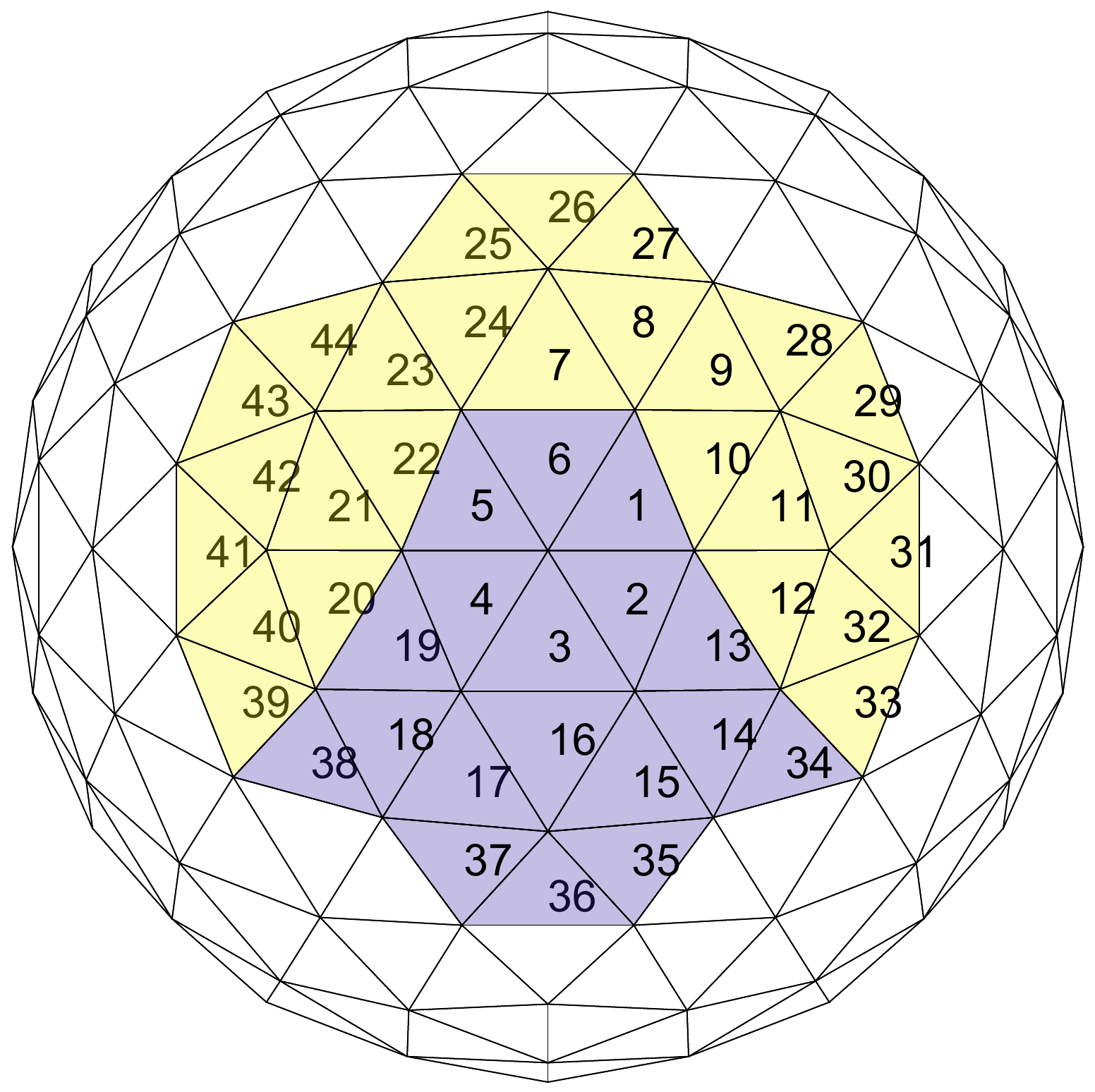}
        \caption{joined hands}
        \label{fig:dynamic5}
\end{subfigure}
\begin{subfigure}[b]{.3\textwidth} \centering
\includegraphics[width=0.7\textwidth]{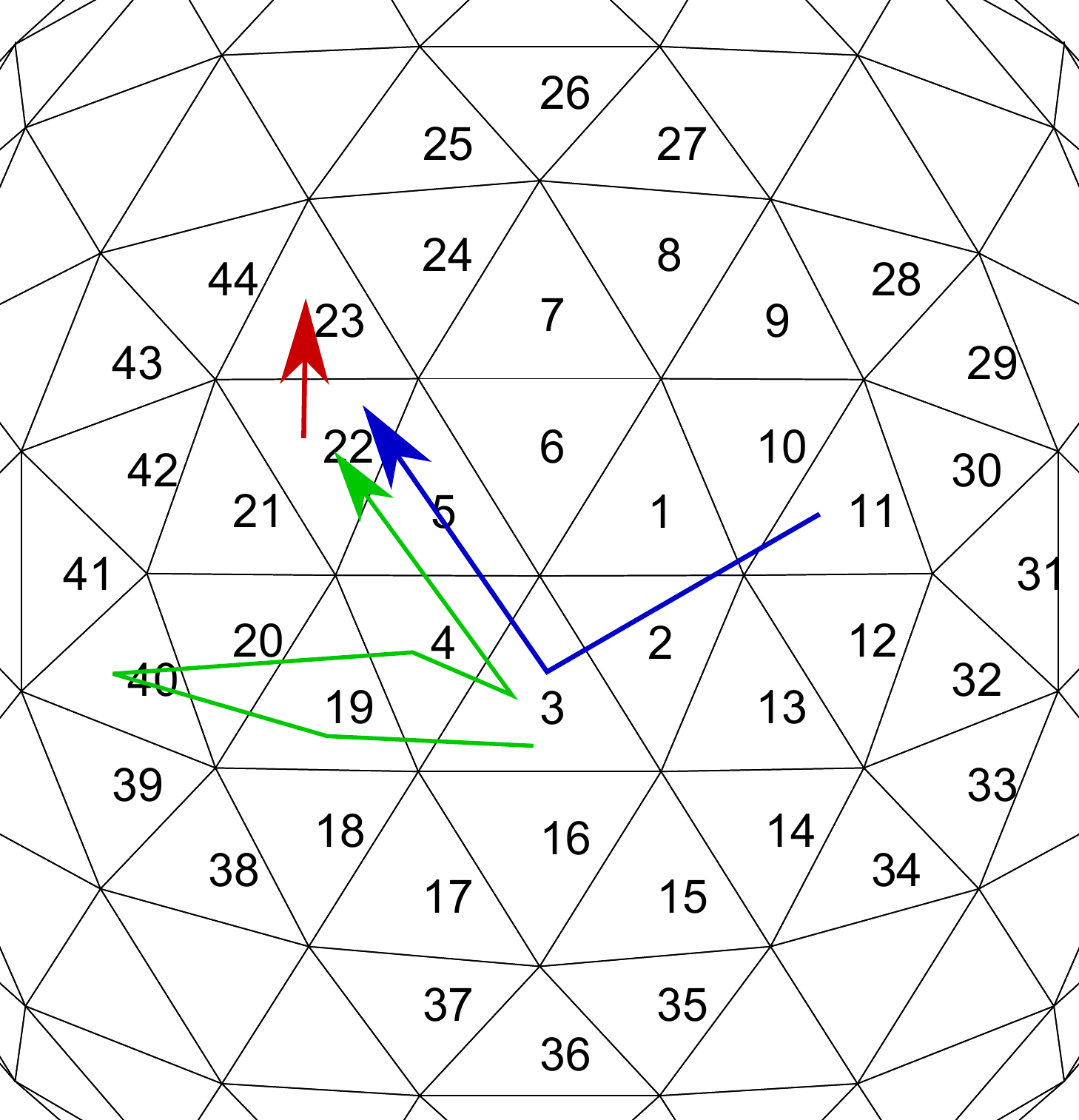}
        \caption{three camera paths}
        \label{fig:dynamic-camera-paths}
\end{subfigure}
\end{center}
  \caption{Dynamic scene with moving obstacles. (a-e) show the occlusion projected on the dome (purple: occluded viewpoints, yellow: not occluded). (f) shows the camera trajectories for different starting positions.}
\label{fig:dynamic}
\end{figure*}

\subsection{Dynamic experiments}
In this section we show the performance of the proposed algorithm in a dynamic scene. 
Since the faces in the dome are large, the occlusion does not change much between consecutive frames. 
Hence, we have chosen 5 key frames where the occlusion changes significantly to see the result of the method.

The occlusion sequence (Fig.~\ref{fig:dynamic}) has an initial position where the views at the top of the dome are occluded (a), then the occlusion moves to the left (b), then another occlusion appears for views on on the other side of the dome (c). Later the two occluded regions go to the top of the dome (d) and finally they merge (e). 
This sequence is the case like an arm covering the top of the target point, moves to one side, then the other arm appears on the other side, and finally both arms move and overlap in the center of the scene.

For this experiment we choose 3 initial positions, corresponding to viewpoints 11 (not occluded), 3 (occluded) and 22 (not occluded). 
The results of the algorithm from the three starting positions (11$\rightarrow$blue, 3$\rightarrow$green, 22$\rightarrow$red) are depicted in Fig.~\ref{fig:dynamic-camera-paths}, which illustrates the sequence of viewpoints given the changing data.
%
%\begin{itemize}
%	\item $ 3 \rightarrow 19 \rightarrow 40 \rightarrow 4 \rightarrow 3 \rightarrow 22$  
%	\item $ 11 \rightarrow 11 \rightarrow 11 \rightarrow 2 \rightarrow 3 \rightarrow 22$  
%	\item $ 22 \rightarrow 22 \rightarrow 23 \rightarrow 23 \rightarrow 23 \rightarrow 23$ 
%\end{itemize}

With a different configuration of parameters where the viewpoint weight is higher, the sequences of movements would have more changes of viewpoint as the algorithm tries to optimize the viewpoint. The dynamic behavior of the proposed method can be also observed in the video presented online\footnote{\url{https://youtu.be/v1vy_FdxCa4}}.

\section{Conclusion}\label{sec:conclusion}

This paper presented both a new problem of dynamically selecting viewpoints for a moving sensor in an environment with static and dynamic obstacles, and provided an algorithm to solve this problem. 
We proposed the use of a set of tessellations of a virtual dome as the set of possible viewpoints. 
The best next viewpoint becomes the one that minimizes four factors, occlusion, quality of viewpoint in terms of angle, geodesic distance and joint distance of the robot arm. 
We have evaluated our proposal in different scenarios showing good performance in all cases.

As future work, we want to explore more parameter configurations which will provide different behavior. 
Moreover, we are interested in including a smoothing factor to avoid rapid changes, and a prediction criterion in order to minimize jumps.

\section*{Acknowledgements}

Authors acknowledge the support of EU project TrimBot2020 and project from University of Alicante (Gre16-28).

{\small
  \bibliographystyle{ieee}
  \bibliography{bibliography}
}

\end{document}